\documentclass{article}

\usepackage{microtype}
\usepackage{graphicx}
\usepackage{subcaption}
\usepackage{booktabs} 

\usepackage{hyperref}

\usepackage[accepted]{icml2024}

\usepackage{amsmath}
\usepackage{amssymb}
\usepackage{mathtools}
\usepackage{amsthm}

\usepackage[capitalize,noabbrev]{cleveref}

\theoremstyle{plain}

\theoremstyle{definition}

\theoremstyle{remark}

\usepackage[textsize=tiny]{todonotes}

\usepackage{amsmath,amsfonts,bm}

\def\Figref#1{Figure~\ref{#1}}

\def\eqref#1{Eq.~(\ref{#1})}
\def\Eqref#1{Eq.~(\ref{#1})}

\def\Algref#1{Algorithm~\ref{#1}}

\def\1{\bm{1}}

\def\vx{{\bm{x}}}

\DeclareMathAlphabet{\mathsfit}{\encodingdefault}{\sfdefault}{m}{sl}
\SetMathAlphabet{\mathsfit}{bold}{\encodingdefault}{\sfdefault}{bx}{n}

\def\gL{{\mathcal{L}}}

\def\gS{{\mathcal{S}}}

\def\gU{{\mathcal{U}}}

\newcommand{\Ls}{\mathcal{L}}

\DeclareMathOperator*{\argmax}{arg\,max}

\DeclareMathOperator{\sign}{sign}

\usepackage{adjustbox}
\usepackage{multirow}

\usepackage{bbold}
\usepackage{physics}
\usepackage{float}

\begin{document}
\twocolumn[
\icmltitle{Downstream Transfer Attack: Adversarial Attacks on Downstream Models with Pre-trained Vision Transformers}



\icmlsetsymbol{equal}{*}

\begin{icmlauthorlist}
\icmlauthor{Weijie zheng}{Fudan University}
\icmlauthor{Xingjun Ma}{Fudan University}
\icmlauthor{Hanxun Huang}{The University of Melbourne}
\icmlauthor{Jun Yin}{Ant Group}
\icmlauthor{Tiehua Zhang}{Tongji University}
\icmlauthor{Zuxuan Wu}{Fudan University}
\icmlauthor{Yu-Gang Jiang}{Fudan University}
\end{icmlauthorlist}

\icmlaffiliation{Fudan University}{Fudan University}
\icmlaffiliation{The University of Melbourne}{The University of Melbourne}
\icmlaffiliation{Ant Group}{Ant Group}
\icmlaffiliation{Tongji University}{Tongji University}

\icmlcorrespondingauthor{Xingjun Ma}{xingjun.ma@unimelb.edu.au}

\icmlkeywords{Machine Learning, ICML}

\vskip 0.3in
]

\begin{abstract}
With the advancement of vision transformers (ViTs) and self-supervised learning (SSL) techniques, pre-trained large ViTs have become the new foundation models for computer vision applications.
However, studies have shown that, like convolutional neural networks (CNNs), ViTs are also susceptible to adversarial attacks, where subtle perturbations in the input can fool the model into making false predictions. This paper studies the transferability of such an adversarial vulnerability from a pre-trained ViT model to downstream tasks.
We focus on \emph{sample-wise} transfer attacks and propose a novel attack method termed \emph{Downstream Transfer Attack (DTA)}. 
For a given test image, DTA leverages a pre-trained ViT model to craft the adversarial example and then applies the adversarial example to attack a fine-tuned version of the model on a downstream dataset. During the attack, DTA identifies and exploits the most vulnerable layers of the pre-trained model guided by a cosine similarity loss to craft highly transferable attacks. Through extensive experiments with pre-trained ViTs by 3 distinct pre-training methods, 3 fine-tuning schemes, and across 10 diverse downstream datasets, we show that DTA achieves an average attack success rate (ASR) exceeding 90\%, surpassing existing methods by a huge margin. When used with adversarial training, the adversarial examples generated by our DTA can significantly improve the model's robustness to different downstream transfer attacks.

\end{abstract}    
\section{Introduction}
\label{sec:intro}

\begin{figure}
  \centering
  \includegraphics[width=.45\textwidth]{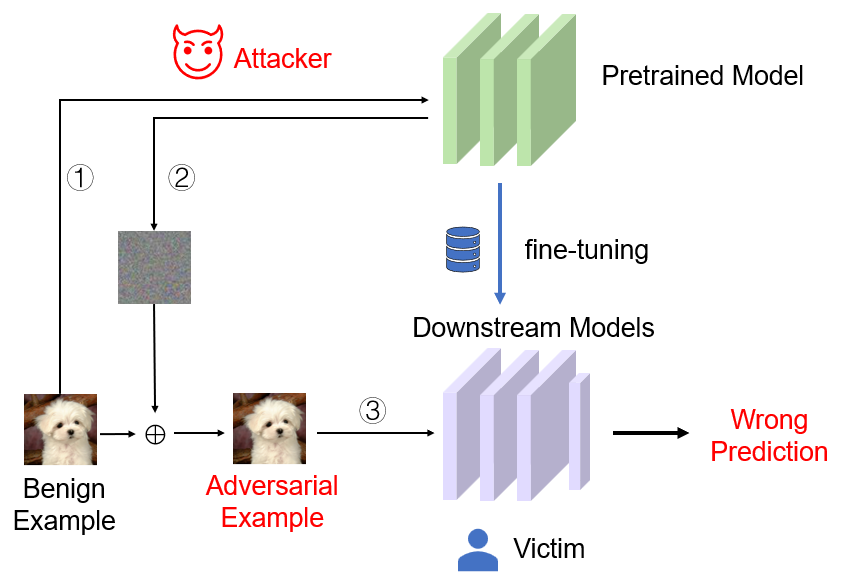}
  \caption{A conceptual illustration of downstream transfer attack.}
  \label{img}
\end{figure}

Due to the exceptional representation capability of large pre-trained models, it has become a common practice to fine-tune a large pre-trained model for a downstream task in both natural language processing (NLP)~\cite{touvron2023llama, brown2020language} and computer vision (CV)~\cite{caron2021dino, steiner2021train, he2022mae} applications. Following this pretraining-and-finetuning paradigm, both traditional supervised learning methods and the emerging self-supervised learning (SSL) methods are devoted to training a generalized representation extractor~\cite{steiner2021train, caron2021dino, oquab2023dinov2, he2022mae}, which can serve as a foundation model for diverse downstream tasks. 
With the help of large pre-trained models, downstream models can be trained with less data and time to obtain a better performance than training from scratch. 
This can be efficiently done by parameter-efficient transfer learning (PETL) methods such as LoRA~\cite{hu2021lora} and AdaptFormer~\cite{chen2022adaptformer}. These methods reduce the number of trainable parameters, accelerating the fine-tuning process, while obtaining on-par or better performance than full fine-tuning~\cite{hu2021lora, chen2022adaptformer, jia2022visual}.

However, deep neural networks (DNNs), including convolutional neural networks (CNNs) and recent vision transformers (ViTs), are vulnerable to adversarial examples, which are slightly perturbed (by an attack method) test inputs that can fool DNNs into making false predictions~\cite{madry2017pgd, carlini2017cw, he2020momentum, moosavi2017uap}. Existing attack methods 
can be categorized into white-box attacks and black-box attacks.
White-box attacks have full access to the model, while black-box attacks can only query the target model to estimate the adversarial gradient~\cite{guo2019simple,chen2017zoo} or leverage a surrogate model to generate transferable adversarial examples~\cite{liu2016delving,dong2019evading,xie2019improving,wei2022towards}. Compared to query-based attacks which often require a large number of expensive queries to the target model, transfer attacks are arguably more easy and cheap to craft against real-world commercial APIs. In this work, we focus on transfer attacks from pre-trained ViTs to fine-tuned downstream models.

Existing transfer attacks are mostly focused on the \emph{cross-model transferability} of adversarial examples, where the surrogate model was trained on the same dataset as the target models~\cite{papernot2016transferability, wei2022towards, dong2018MIM, zhang2023transferable, wu2020skip-connection}. However, under the current pretrain-and-finetune paradigm, the downstream models are often fine-tuned on a dataset or task that is completely different from the pre-training. We call this setting \emph{downstream transfer}, which can be regarded as one special case of \emph{cross-domain transfer} attacks~\cite{naseer2019crossdomain,zhang2022BIA}. Under the downstream transfer setting, the adversary generates (untargeted) adversarial examples using the pre-trained model and then applies the adversarial examples to attack a fine-tuned version of the model on a downstream dataset. This setting has recently been studied in two research works~\cite{ban2022pap, zhou2023downstream_agnostic} from the perspective of universal adversarial perturbations (UAPs). Different from these works, we focus on \emph{sample-wise} downstream transfer attacks and pre-trained ViTs.

Particularly, we propose a new attack method named \textbf{\emph{Downstream Transfer Attack}} (DTA) to generate highly transferable adversarial attacks from pre-trained ViTs to downstream models. DTA iteratively minimizes the average cosine similarity between the tokens of clean and adversarial examples at the most vulnerable layer(s) of the pre-trained model. The cosine similarity is used as the adversarial objective as well as an indicator of the transfer strength of the attack.
DTA first targets the shallow layers of the model, as low-level features are more amenable to transfer.   If this initial attempt is unsuccessful, i.e., the cosine similarity is still above a certain threshold, DTA probes all intermediate layers to identify the most vulnerable layer(s) that minimizes the cosine similarity. It then combines all these layers to perform the final attack.

In summary, our main contributions are:
\begin{itemize}
    \item We investigate the adversarial vulnerability of the current pretrain-and-finetune paradigm and propose a new attack method named \emph{Downstream Transfer Attack} (DTA) to generate transferable attacks from pre-trained ViTs to downstream models. 

    \item We empirically find that the cosine similarity between clean and adversarial tokens is a good indicator of the downstream transferability and leverage this finding to identify the most vulnerable (and transferable) layer(s) of a pre-trained ViT in DTA. Extensive experiments demonstrate the strong performance of our proposed DTA attacks. The results also show that, compared to full fine-tuning, typical PETL methods introduce more vulnerability to downstream models.
    
    \item We also explore the possibility of using DTA to improve adversarial training based defense and find that, compared to UAP attack, adversarial examples generated by our sample-wise attack DTA can significantly improve the model's robustness to different downstream transfer attacks.
\end{itemize}
\section{Related Work}
\label{sec:related work}
\paragraph{Pre-training and Fine-tuning}
Pre-training a large model on a large-scale dataset equips the model with a foundational ability to extract all levels of features from the input. Existing pre-training methods can be roughly classified into supervised learning~\cite{khosla2020supervised, Krizhevsky2012imagenet, steiner2021train} and self-supervised learning (SSL) methods. SSL allows the model to learn directly from web-scale data without label annotations, thus becoming a popular and practical choice for large-scale pre-training. SSL methods can be further broadly categorized into contrastive methods~\cite{Garrido2022Ontheduality, caron2021dino, chen2020simple, he2020momentum, grill2020bootstrap} and generative methods~\cite{he2022mae,bao2021beit}. 
\vspace{+2mm}

Fine-tuning adapts a pre-trained model to a specific downstream task by training on the downstream dataset. As the pre-trained models become larger and larger, traditional full fine-tune tends to suffer from efficiency and storage limitations. This motivates the proposal of parameter-efficient transfer learning (PETL) methods~\cite{chen2022adaptformer, hu2021lora,jia2022visual}. These methods involve freezing the weights of the pre-trained model and introducing auxiliary fine-tunable modules, thus having the ability to achieve comparable performance with full fine-tuning while saving a lot of parameter updates and storage. For instance, LoRA leverages low-rank matrices to represent the updates of attention block parameters~\cite{hu2021lora}, while AdaptFormer attaches parallel adapters to the fully connected layers ~\cite{chen2022adaptformer}.

\paragraph{Transferable Adversarial Attacks}
Transferable adversarial attacks, or transfer attacks for short, are a form of black-box attacks that leverage the cross-model transferability of adversarial examples~\cite{liu2016delving}.
Most of the existing works were focused on the cross-model transferability, where the adversary generates the attack using a surrogate model trained on the same training dataset as the target model~\cite{dong2018MIM, dong2019evading, xie2019improving, wu2020skip-connection, zhang2023transferable, wei2022towards, t1, t2, t3}. The transfer techniques used by these methods involve the labels or intermediate layer feature maps of the network. For example, Wang et al.~\cite{wang2021feature-importance-aware} and Zhang et al.~\cite{zhang2022improving} employ intermediate feature maps to identify the important regions in the image to improve transferability. There are also methods that directly perturb the intermediate layer features to craft transferable adversarial examples~\cite{naseer2018NRDM, ganeshan2019fda}.

Universal Adversarial Perturbations (UAPs)~\cite{moosavi2017uap, khrulkov2018uap-art, weng2023uap-comparative, mopuri2017uap-fastfeaturefool} can also be viewed as one type of cross-sample transfer attacks, where a single adversarial pattern works for different images. UAPs are often generated by accumulating the adversarial perturbations over different samples~\cite{moosavi2017uap, mopuri2017uap-fastfeaturefool}. Moreover, UAPs can further be combined with other types of transferability like cross-dataset (task) transferability to generate UAPs that transfer to downstream models~\cite{ban2022pap, zhang2021datafreeuap}.

\paragraph{Downstream Transfer Attacks} The downstream transfer attack setting from a pre-trained model to downstream models has been studied before. But these works were mainly focused on UAPs. For example, Ban and Dong~\cite{ban2022pap} proposed Pre-trained Adversarial Perturbation (PAP) to attack different downstream models using UAPs generated based on the pre-trained model. Specifically, the UAPs were generated by maximizing the $L_2$ norm of the model's shallow layer features based on the pre-training dataset. Alternatively, a recently proposed AdvEncoder~\cite{zhou2023downstream_agnostic} attack adopts a generative approach to craft downstream-agnostic UAPs. It trains the UAP generator using an adversarial objective that consists of three key components: a high-frequency component loss, a quality loss, and an adversarial loss. 
\vspace{+2mm}

Meanwhile, traditional transfer attacks that are purely feature-based can also be applied here to attack downstream models. For example, the Feature Disruptive Attack (FDA)~\cite{ganeshan2019fda} which diminishes activations that support the current prediction while strengthening activations that do not,  and Neural Representation Distortion Method (NRDM)~\cite{naseer2018NRDM} attack which directly maximizes the perceptual metric defined on feature maps.
It is worth mentioning that downstream transfer can also occur between a pre-trained multimodal model like CLIP and its downstreams~\cite{cjj2023downstream, zhou2023advclip}. In this paper, we focus on the typical downstream transfer setting introduced in PAP~\cite{ban2022pap}, where the transfer occurs from a  ViT pre-trained on vision dataset to a downstream classifier. But different from PAP, we focus on sample-wise transfer attacks rather than UAPs.

\section{Downstream Transfer Attack}
\label{sec:DTA}

\paragraph{Notations}
Let $f_{\theta}$ be the pre-trained encoder and $f^{k}_{\bm{\theta}}(\vx)$ is its $k$-th layer feature map output for a given input image $\vx$. Let $f_{\theta'}$ be the model fine-tuned from $f_{\theta}$ on a downstream dataset $\mathcal{D}_d$. Let $\vx\sim\mathcal{D}_{d}$ be an image in dataset $\mathcal{D}_{d}$, and $f_{\theta'}(\vx)$ be the probability output of classifier $f_{\theta'}$. Let ${F}_{\theta'}(\vx) = \argmax f_{\theta'}(\vx)$ be the final classification result.

\subsection{Threat Model}
In our downstream transfer attack setting, the adversary aims to attack a target model $f_{\bm{\theta'}}$ which was fine-tuned from a pre-trained model $f_{\bm{\theta}}$. The fine-tuning was done by either full fine-tuning or a PETL method on a downstream dataset that is different from the pre-training dataset. The adversary has no knowledge or access to the fine-tuning process and can only query the target model to mount the attack. However, the adversary has full access to the pre-trained model which is often assumed to be a large open source model. Given a test image $\vx$ of the downstream dataset, the adversary exploits the pre-trained model to generate the adversarial perturbation $\delta$ with attack budget $\norm{\delta}_p \leq \epsilon$. The resulting adversarial example $\vx' = \vx+\delta$ is then fed into the downstream model $f_{\theta'}$ to execute the attack.

The attacker's goal is to maximize the loss of the downstream model $f_{\theta'}$:
\begin{equation}
\begin{aligned}
\max \Ls (f_{\theta'}(\vx+\delta)),\;
s.t. \norm{\delta}_p \leq \epsilon,
\end{aligned}
\label{eq:downstream attack}
\end{equation}

where $\Ls$ denotes the loss of the downstream task. In the case of image classification task, the above adversarial objective can be defined as the misclassification error:

\begin{equation}
\max \mathbb{1}( F_{\theta'}(\vx+\delta) \neq y),\;
s.t. \norm{\delta}_p \leq \epsilon,
\label{eq:downstream attack img clf}
\end{equation}
where $\mathbb{1}(\cdot)$ denotes the indicator function and $y$ denotes the ground truth label.

\paragraph{Relation to Existing Threat Models} Our downstream transfer threat model is an extension of the threat model proposed in PAP~\cite{ban2022pap} and AdvEncoder~\cite{zhou2023downstream_agnostic}, which assume that the pre-trained model is known to the adversary. In our threat model, the adversary can access both the pre-trained model and the test samples that the attacker wishes to attack, 
but does not know the fine-tuned parameters or architecture of the downstream model. 
This allows us to generate sample-wise transfer attacks, while PAP and AdvEncoder can only generate UAPs (as the adversary has no knowledge of the downstream dataset). 
\textbf{The advantages of sample-wise transfer attacks over UAPs are twofold: 1) they are stronger than UAPs, revealing more severe threats; and 2) they can help train more robustness models when used for adversarial training.}
Our attack setting can also be viewed as a special case of cross-domain transfer attacks~\cite{naseer2019crossdomain,zhang2022BIA}, but with a special correlation between the source and target models, i.e., the target model was fine-tuned from the source model. Our setting is also related to cross-modality transfer attacks~\cite{cjj2023downstream,zhou2023advclip}, which use UAPs generated using CLIP~\cite{radford2021learning} to attack downstream tasks like image-text retrieval and image classification. Our work focuses on the transfer from ViTs pre-trained on large-scale datasets to downstream tasks including image classification, object detection, and semantic segmentation.

\subsection{Methodology}

Arguably, the most crucial part of designing a transferable attack is to find an appropriate indicator of transferability. This has been found to be challenging in the traditional cross-model transfer setting~\cite{liu2016delving,dong2018MIM,dong2019evading}. This is because strong attacks generated on a source model tend to overfit the source model and thus transfer poorly to other (target) models, even if the target models are trained on the same dataset~\cite{lu2020enhancing, zhang2022improving, zhang2023transferable}. We find that this is also the case for downstream transfer attacks where the fine-tuned models often have substantial parameter changes caused by either full finetuning, self-attention adaptation (e.g., LoRA), or additional adapter layers (e.g., AdaptFormer). In this work, we propose a novel metric called \emph{Average Token Cosine Similarity (ATCS) } as the indicator of downstream transferability and also our adversarial objective. Based on ATCS, we further introduce a layer selection strategy to find the most vulnerable and transferable layer of a pre-trained ViT to downstream models.

\begin{algorithm}[tb]
    \caption{Downstream Transfer Attack (DTA)}
    \label{alg:DTA}
\begin{algorithmic}
    \STATE {\bfseries Input:} pre-trained model $f_{\theta}$, clean example $\vx$, attack budget $\epsilon$, threshold $\gamma$, the total number of layers $M$, the number of attacked layers $N$
    
    \STATE {\bfseries Output:} adversarial example $\vx^{*}$
    
    \STATE initialize $ \vx^{*} = \vx + \gU(-\epsilon, \epsilon)$
    
    \STATE update $\vx^{*}$ by \Eqref{eq:optimize} with $\Ls = \Ls_{\text{ATCS}}^{1} (\vx, \vx^{*})$, denote the final loss as $l$ 
    \IF{$l < \gamma$}
        \STATE return $\vx^{*}$
    \ELSE
        \STATE initialize $ \vx' = \vx + \gU(-\epsilon, \epsilon)$ \\
        \STATE let $M' = \{\lfloor M/3 \rfloor, \lfloor M/3\rfloor+1, \cdots, M\}$\\
        \STATE update $\vx'$ by \eqref{eq:optimize} with $\Ls = \sum_{i \in M'} \Ls_{\text{ATCS}}^{i}(\vx,\vx')$ \\
        \STATE calculate $\gS_{\gL} = \{\Ls_{\text{ATCS}}^{i}(\vx,\vx')| i \in M'\}$\\
        \STATE denote the $N$-th smallest element in $\gS_{\gL}$ as $\Ls_{\text{ATCS}}^{*}$\\
        \STATE $M^*=\{m \in M'| \Ls_{\text{ATCS}}^{m}(\vx,\vx') \leq \Ls_{\text{ATCS}}^{*}\}$\\
        \STATE let $ \vx^{*} = \vx + \gU(-\epsilon, \epsilon)$ \\
        \STATE update $\vx^{*}$ by \eqref{eq:optimize} with $\Ls = \sum_{i \in M^*}\Ls_{\text{ATCS}}^{i}(\vx,\vx^{*})$ \\
        \STATE return $\vx^{*}$ 
    \ENDIF
\end{algorithmic}
\end{algorithm}

\paragraph{Average Token Cosine Similarity}
Since pre-trained models do not necessarily have a classification head, the adversarial objective should be defined on the intermediate layer output (i.e., features) of the pre-trained model. A typical ViT model consists of a series of identical transformer layers (blocks). 
The output of a transformer layer is a sequence of tokens that have the same dimensions. 
As such, we propose to first compute the cosine similarity between the clean vs. adversarial feature tokens and then average this cosine similarity across the tokens to obtain the final \emph{ATCS}. 
Formally, the ATCS loss is defined as:
\begin{equation}
\Ls_{\text{ATCS}}^{k}(\vx, \vx') = \frac{1}{|T|} \sum_{t \in T}{cos(f^{k}_{\bm{\theta}}(\vx)_{t},f^{k}_{\bm{\theta}}(\vx')_{t})}, 
\label{eq:DTA}
\end{equation}
where $f^{k}_{\bm{\theta}}(\vx)$ is the feature map output of the $k$-th layer,
$f^{k}_{\bm{\theta}}(\vx)_t$ is the $t$-th token of $f^{k}_{\bm{\theta}}(\vx)$, and $|T|$ is the total number of tokens.

The reason why we use ATCS over the traditional cosine similarity computed on flattened tokens is that the high dimensionality of the transformer layer output could cause the traditional cosine similarity between different samples to converge to the same value, which is known as the curse of dimensionality. For example, a transformer layer could have 197 tokens, with each token of 768 dimensions, adding up to a 151296 dimension vector once flattened. The use of ATCS could mitigate the curse of dimensionality, as the cosine similarity is only computed on 768-dimensional vectors.

With the ATCS loss, we can generate an adversarial example iteratively as follows:
\begin{equation}
\begin{aligned}\label{eq:optimize}
\vx'_0 &= \vx + \gU(-\epsilon,\epsilon) \\
\vx'_{t+1} &= \text{Clip}_{\vx,\epsilon}\left\{\vx'_{t} + \eta \cdot \sign( \nabla \Ls_{\text{ATCS}}^{k}(\vx, \vx'_{t}))\right\},\\
\end{aligned}
\end{equation}
where, $\vx$ is a clean sample, $\vx'_{t}$ is the adversarial example obtained at perturbation step $t$, $\epsilon$ is the maximum perturbation ($\epsilon=10/255$), $\text{Clip}_{\vx,\epsilon}$ is a clip operation that clips the perturbed sample to be within the $\epsilon$-ball around $\vx$, $\gU(-\epsilon,\epsilon)$ is a uniform noise added to $\vx$ as an initialization, and $\eta$ is the perturbation step size. The above formulation follows the classic Projected Gradient Descent (PGD)~\cite{madry2017pgd} attack.

\begin{figure}
\centering
\includegraphics[width=.45\textwidth]{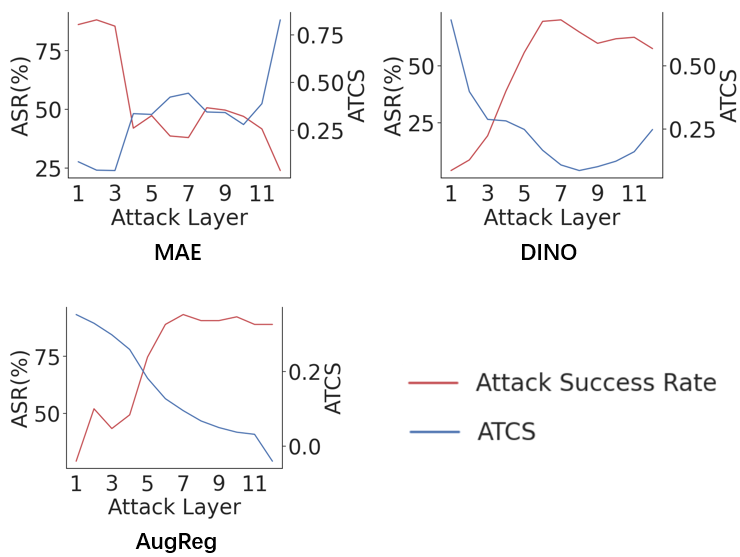} 

\caption{ATCS vs. downstream transfer Attack Success Rate (ASR). The source ViT-base models were pre-trained by MAE, DINO, and AugReg, while their downstream (target) models were fully finetuned on CIFAR-10.}
\label{fig:cos_and_asr}
\end{figure}

\begin{figure*}
\centering
	\subcaptionbox{MAE\label{subfig:mae}}{\includegraphics[width = 0.32\textwidth]{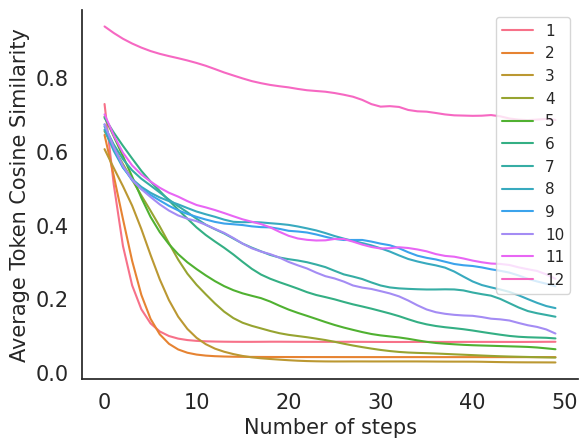}}
	\subcaptionbox{DINO\label{subfig:dino}}{\includegraphics[width = 0.32\textwidth]{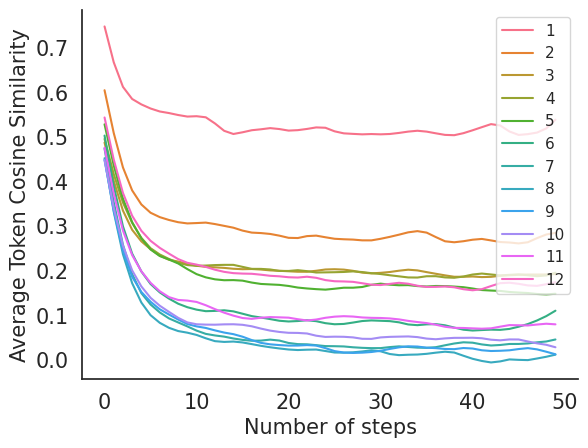}}
	\subcaptionbox{AugReg\label{subfig:augreg}}{\includegraphics[width = 0.32\textwidth]{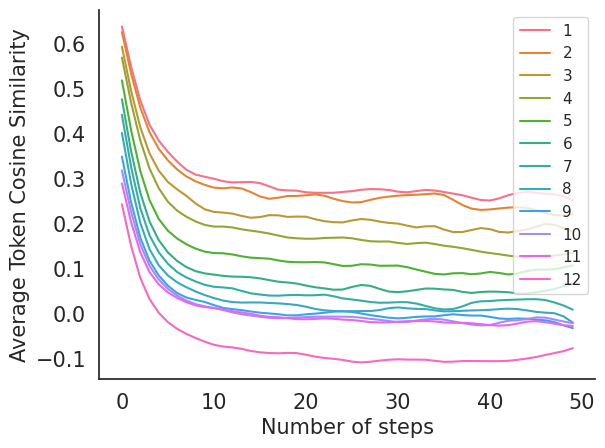}} 

\caption{The ATCS values of adversarial examples generated at different steps of \Eqref{eq:optimize} for CIFAR-10 test images. Each line indicates attacking a particular layer of the pre-trained model. Figure \subref{subfig:mae}, \subref{subfig:dino}, and \subref{subfig:augreg} represent the source ViT-base models pre-trained by MAE, DINO, and AugReg on ImageNet, respectively. Their downstream models were fully fine-tuned on CIFAR-10.}
\label{fig:cos_pretrain}
\end{figure*}

The above attack generation formulation does not solve the transferability problem and we still need a good transferability indicator. Here, we show that the ATCS loss value itself is a good and consistent indicator of transferability. Specifically, we will show that a lower ATCS value leads to better transferability (measured by the attack success rate on the downstream model), regardless of which layer the value is obtained.
We consider 3 pre-trained ViT-base models by MAE\cite{he2022mae}, DINO\cite{caron2021dino}, and AugReg\cite{steiner2021train} on the ImageNet dataset as the source models, and their full fine-tuned version on the CIFAR-10 dataset as the downstream (target) models. Figure \ref{fig:cos_and_asr} illustrates the negative correlation between the ATCS values achieved by our attack (defined in \eqref{eq:optimize}) at different layers of the pre-trained model and the downstream transfer attacks success rate: \emph{the transfer ASR rises whenever the ATCS decreases}. This validates that ATCS is a good indicator of downstream transferability. 


\paragraph{Finding the Most Vulnerable Layers}
With ATCS, the next step is to find the most vulnerable layer of the pre-trained model that could lead to the lowest ATCS value. \Figref{fig:cos_pretrain} shows that the most vulnerable (transferable) layer of the source models pre-trained by different methods (i.e., MAE, DINO, and AugReg) are quite different. Particularly, for the MAE pre-trained model, shallower layers (1, 2, 3) generally lead to faster reduction in ATCS and lower ATCS values in the end. However, for the DINO pre-trained model, the middle layers (7, 8, 9) produce the lowest ATCS, and the most vulnerable layers of the AugReg pre-trained model are the deep layers (11, 12).
Next, we will introduce our method of finding the most vulnerable layer for different pre-trained models. 

An intuitive idea to find the most vulnerable layer is to attack each layer of the pre-trained model separately. The adversary can then check the achieved ATCS value and take the adversarial examples with the lowest ATCS as the final attacks. However, the complexity of such an algorithm increases linearly with the number of layers $M$ and could become extremely slow when $M$ is a large number. Moreover, attacking the individual layers separately could also miss important vulnerabilities of combined layers. Fortunately, this problem can be easily addressed, as we find that attacking the middle and deep layers all at once yields a similar result (ATCS ranking) as attacking them individually. Motivated by this observation, we propose the following layer selection strategy. 

First, we attack the shallowest (i.e., first) layer of the model and check the achieved ATCS. If the ATCS is below a certain threshold $\gamma$, the attack is considered successful, i.e., there is no need to further attack the middle or deep layers. Otherwise, we continue to attack the $\lfloor M/3 \rfloor \sim M$-th layer all together using the combined loss function $\sum_{k=\lfloor M/3 \rfloor}^{M}\Ls^{k}_{\text{ATCS}}$. We then rank these layers according to their ATCS value indicated by $\Ls^{k}_{\text{ATCS}}$. The top-$N$ layers with the lowest ATCS value will be selected as the most vulnerable layers, which will be attacked again to obtain the final adversarial examples. One could also assign different weights to the layers, however, we find it does not necessarily improve the attack performance. 
The complete algorithm of our attack is described in \Algref{alg:DTA}.

\section{Experiments}
\label{sec:experments}

\begin{table*}[htbp]
  \centering
  \caption{The ASR (\%) of different attack methods on downstream ViT-base models \textbf{full fine-tuned} from different pre-trained models. \textbf{Avg. 1} averages over the datasets, while \textbf{Avg. 2} averages over the datasets and pre-training methods.}
  \begin{adjustbox}{width=0.98\linewidth}
    \begin{tabular}{c|c|cccccccccc|c|c}
    \toprule
    Attack & \multicolumn{1}{c}{Pretrain} & CIFAR10 & CIFAR100 & STL10 & Cars  & Cub   & DTD   & FGVC  & Food  & Pets  & SVHN  & Avg. 1 & Avg. 2 \\
    \midrule
    \multirow{3}[2]{*}{NRDM} & AugReg & 87.47  & 98.28  & 89.49  & 98.16  & 98.02  & 92.50  & 97.24  & 98.60  & 96.02  & 80.24  & 93.60  & \multirow{3}[2]{*}{82.49 } \\
          & MAE   & 33.14  & 79.34  & 65.79  & 80.60  & 72.61  & 83.35  & 84.61  & 90.31  & 80.21  & 44.15  & 71.41  &  \\
          & DINO  & 68.35  & 96.50  & 87.38  & 92.46  & 90.73  & 88.56  & 94.09  & 97.52  & 95.45  & 13.66  & 82.47  &  \\
    \midrule
    \multirow{3}[2]{*}{PAP} & AugReg & 19.71  & 42.06  & 5.37  & 34.10  & 37.22  & 56.38  & 56.16  & 48.63  & 13.60  & 30.71  & 34.39  & \multirow{3}[2]{*}{50.72 } \\
          & MAE   & 89.98  & 98.69  & 89.86  & 99.42  & 99.32  & 94.84  & 99.00  & 98.92  & 97.02  & 80.30  & 94.74  &  \\
          & DINO  & 8.43  & 47.08  & 4.17  & 17.11  & 28.14  & 38.45  & 49.71  & 21.73  & 10.02  & 5.38  & 23.02  &  \\
    \midrule
    \multirow{3}[2]{*}{\textbf{DTA}} & AugReg & 91.94  & 99.06  & 93.33  & 99.65  & 98.50  & 96.12  & 99.07  & 99.71  & 97.14  & 89.88  & 96.44  & \multirow{3}[2]{*}{\textbf{93.11 }} \\
          & MAE   & 89.92  & 98.80  & 89.48  & 99.29  & 98.79  & 92.45  & 99.07  & 98.97  & 96.65  & 84.27  & 94.77  &  \\
          & DINO  & 74.82  & 99.25  & 89.35  & 97.18  & 97.62  & 93.99  & 97.84  & 98.91  & 95.26  & 37.09  & 88.13  &  \\
    \bottomrule
    \end{tabular}%
    \end{adjustbox}
  \label{tab:full}%
\end{table*}%

\begin{table*}[htbp]
  \centering
  \caption{The ASR (\%) of different attacks on downstream ViT-base models fine-tuned by \textbf{LoRA}.  \textbf{Avg. 1} and \textbf{Avg. 2} are the same as in Table \ref{tab:full}.}
  \begin{adjustbox}{width=0.98\linewidth}
    \begin{tabular}{c|c|cccccccccc|c|c}
    \toprule
    Attack & \multicolumn{1}{c}{Pretrain} & CIFAR10 & CIFAR100 & STL10 & Cars  & Cub   & DTD   & FGVC  & Food  & Pets  & SVHN  & Avg. 1 & Avg. 2 \\
    \midrule
    \multirow{3}[2]{*}{NRDM} & AugReg & 87.50  & 96.76  & 88.70  & 97.82  & 98.26  & 92.55  & 96.88  & 98.34  & 96.10  & 77.29  & 93.02  & \multirow{3}[2]{*}{85.81 } \\
          & MAE   & 54.34  & 81.83  & 83.44  & 64.97  & 69.26  & 81.22  & 78.73  & 90.55  & 75.99  & 50.05  & 73.04  &  \\
          & DINO  & 88.61  & 97.85  & 89.63  & 97.11  & 97.24  & 91.06  & 94.90  & 97.85  & 96.97  & 62.64  & 91.39  &  \\
    \midrule
    \multirow{3}[1]{*}{PAP} & AugReg & 30.47  & 43.87  & 6.23  & 42.29  & 55.03  & 54.30  & 65.28  & 53.65  & 19.92  & 85.21  & 45.63  & \multirow{3}[1]{*}{63.22 } \\
          & MAE   & 90.17  & 99.06  & 90.00  & 99.39  & 99.39  & 96.22  & 99.00  & 99.01  & 97.27  & 84.43  & 95.39  &  \\
          & DINO  & 60.39  & 88.84  & 7.93  & 33.14  & 48.46  & 43.13  & 59.16  & 42.52  & 15.50  & 87.40  & 48.65  &  \\
    \midrule
    \multirow{3}[1]{*}{\textbf{DTA}} & AugReg & 93.69  & 99.19  & 92.28  & 99.33  & 99.15  & 96.65  & 99.01  & 99.53  & 97.36  & 81.62  & 95.78  & \multirow{3}[1]{*}{\textbf{95.22 }} \\
          & MAE   & 90.52  & 99.10  & 89.96  & 99.29  & 98.83  & 92.29  & 98.86  & 99.01  & 97.30  & 90.53  & 95.57  &  \\
          & DINO  & 90.31  & 99.62  & 88.49  & 98.69  & 98.98  & 97.07  & 97.90  & 98.83  & 95.34  & 78.00  & 94.32  &  \\
    \bottomrule
    \end{tabular}%
    \end{adjustbox}
  \label{tab:lora}%
\end{table*}%

\begin{table*}[htbp]
  \centering
  \caption{The ASR (\%) of different attacks on downstream ViT-base models tuned by \textbf{AdaptFormer}.  \textbf{Avg. 1} and \textbf{Avg. 2} are the same as in Table \ref{tab:full}.}
  \begin{adjustbox}{width=0.98\linewidth}
    \begin{tabular}{c|c|cccccccccc|c|c}
    \toprule
    Attack & \multicolumn{1}{c}{Pretrain} & CIFAR10 & CIFAR100 & STL10 & Cars  & Cub   & DTD   & FGVC  & Food  & Pets  & SVHN  & Avg. 1 & Avg. 2 \\
    \midrule
    \multirow{3}[2]{*}{NRDM} & AugReg & 86.70  & 98.76  & 89.86  & 98.64  & 98.52  & 94.47  & 97.33  & 98.32  & 96.57  & 66.83  & 92.60  & \multirow{3}[2]{*}{92.53 } \\
          & MAE   & 77.39  & 95.19  & 88.10  & 98.20  & 96.91  & 89.95  & 97.69  & 95.90  & 96.38  & 68.05  & 90.38  &  \\
          & DINO  & 89.12  & 98.46  & 90.06  & 99.14  & 98.31  & 91.97  & 98.08  & 98.51  & 97.36  & 85.16  & 94.62  &  \\
    \midrule
    \multirow{3}[2]{*}{PAP} & AugReg & 31.63  & 59.87  & 7.13  & 52.07  & 59.26  & 59.20  & 68.25  & 58.01  & 19.95  & 62.94  & 47.83  & \multirow{3}[2]{*}{66.50 } \\
          & MAE   & 90.00  & 98.43  & 88.13  & 99.50  & 99.48  & 97.39  & 99.00  & 99.01  & 97.27  & 81.93  & 95.01  &  \\
          & DINO  & 68.76  & 90.29  & 16.32  & 59.07  & 62.02  & 44.68  & 76.62  & 44.11  & 19.37  & 85.28  & 56.65  &  \\
    \midrule
    \multirow{3}[2]{*}{\textbf{DTA}} & AugReg & 90.13  & 99.25  & 92.86  & 99.65  & 98.88  & 97.34  & 99.22  & 99.50  & 97.77  & 76.88  & 95.15  & \multirow{3}[2]{*}{\textbf{95.32 }} \\
          & MAE   & 89.98  & 98.36  & 89.60  & 99.41  & 99.46  & 96.44  & 99.01  & 99.01  & 97.30  & 80.45  & 94.90  &  \\
          & DINO  & 90.93  & 99.37  & 89.49  & 99.20  & 99.21  & 97.39  & 98.98  & 98.91  & 95.20  & 90.28  & 95.90  &  \\
    \bottomrule
    \end{tabular}%
    \end{adjustbox}
  \label{tab:adapter}%
\end{table*}%

\subsection{Experimental Setup}
\paragraph{Models and Datasets} We consider models pre-trained by three representative pre-training methods: AugReg~\cite{steiner2021train} (a supervised pre-training method), DINO~\cite{caron2021dino} (a contrastive learning method), and MAE~\cite{he2022mae} (a masked image modeling method). We utilize the public model weights pre-trained by the three methods on GitHub or Hugging Face. We consider three representative fine-tuning approaches: full fine-tune, LoRA~\cite{hu2021lora}, and AdaptFormer~\cite{chen2022adaptformer}. Following~\cite{ban2022pap}, we use 10 downstream datasets to evaluate the performance of DTA. These datasets contain 3 coarse-grained and 7 fine-grained datasets.
The images are resized to 256 × 256 and then center-cropped to 224 × 224 before feeding into the network. The clean performances of the models are provided in the appendix.

\paragraph{Baseline Methods}
We take PAP~\cite{ban2022pap} and NRDM~\cite{naseer2018NRDM} as our baselines. PAP, which generates image-agnostic perturbation, is the first method for attacking fine-tuned models. NRDM is a sample-wise attack initially introduced for black-box attacks on diverse tasks. Since NRDM operates on features, it can be directly applied to the downstream attack setting. We empirically set the attacking layer of NRDM to $k=8$ for ViT small and base, and $k=16$ for ViT-large, as they yield the best performance.

\paragraph{Attack Setting}
Following previous studies~\cite{zhou2023downstream_agnostic, moosavi2017uap, zhang2022BIA}, we focus on $l_{\infty}$ norm adversarial attack and set the attack budget to $\epsilon = 10/255$. We use 3 steps with step size $\eta = 0.05$ for attacking the first layer and 20 steps with $\eta = 0.02$ for the intermediate layers, as we find that a relatively larger step size with fewer steps performs better at the shallow layers. We set the threshold to $\gamma= 0.25$ and the number of attack layers to $N = 4$. We tuned the hyper-parameters and reported the best result for all attack methods.

\paragraph{Evaluation Metric}
We mainly focus on image classification tasks. Following~\cite{ban2022pap}, we use Attack Success Rate (ASR) as the evaluation metric, which is defined as the proportion of classification errors made by the fine-tuned model over the entire downstream test dataset. We also test the transferability to downstream object detection and segmentation tasks. For object detection, we use mean Average Precision (mAP) as the evaluation metric, while for segmentation, we use mean Intersection over Union (mIoU).

\subsection{Main Results}
For downstream transfer attacks, it is important to achieve good attack performance across different fine-tuning methods, pre-training methods, and downstream datasets. A quick glance at the results in Table \ref{tab:full}, \ref{tab:lora}, and \ref{tab:adapter} reveals that our DTA surpasses the baselines by a huge margin across all pre-training methods and fine-tuning datasets. For full fine-tuning (Table \ref{tab:full}), DTA achieves an average ASR of 93.11\%, surpassing that of NRDM and PAP by more than $10\%$ and $40\%$, respectively. A similar result is also observed for LoRA (Table \ref{tab:lora}) and AdaptFormer (Table \ref{tab:adapter}), where DTA achieves an ASR above 95\%. Comparing the results between full fine-tune and LoRA/AdaptFormer, we find that PETL fine-tuned models (by LoRA/AdaptFormer) are more vulnerable to downstream transfer attacks than full fine-tuned models. This is because the parameters of the pre-trained models are all fixed in PETL, leaving more feature vulnerability to the downstream models. This also confirms that PETL makes less feature change to the pre-trained model than full fine-tuning.

Another interesting observation is that the baselines exhibit a much higher variance when applied to different pre-trained models, and even fail in certain cases. For example, PAP works pretty well on MAE pre-trained models, yet fails badly on AugReg and DINO pre-trained models. It is quite the opposite for NRDM, which works worse on MAE pre-trained models. We conjecture this is because MAE focuses more on reconstruction, making shallow layer features fundamental for the gradual reconstruction at the deep layers. For pre-training methods that focus more on the clustering effect of the deep features, e.g., AugReg and DINO, they are more vulnerable at the middle or deep layers (see \Figref{fig:layer} in the appendix).

\subsection{Evaluation on Large-Scale ViTs}

\begin{table}[htbp]
  \centering
  \caption{The ASR (\%) of different attacks on models of different scales. All models are pre-trained by AugReg and then fully fine-tuned. \textbf{Avg. 1} averages over the datasets, while \textbf{Avg. 2}  averages over the datasets and models.}
    \begin{adjustbox}{width=0.98\linewidth}
    \begin{tabular}{c|c|cccc|c|c}
    \toprule
    Attack & Model & CIFAR10 & CIFAR100 & Food  & SVHN  & Avg. 1 & Avg. 2 \\
    \midrule
    \multirow{3}[2]{*}{NRDM} & ViT-Small & 74.55  & 93.44  & 85.74  & 43.89  & 74.41  & \multirow{3}[2]{*}{74.60 } \\
          & ViT-Base & 87.47  & 98.28  & 98.60  & 80.24  & 91.15  &  \\
          & ViT-Large & 51.56  & 78.01  & 87.58  & 15.78  & 58.23  &  \\
    \midrule
    \multirow{3}[2]{*}{PAP} & ViT-Small & 4.53  & 22.62  & 28.93  & 11.03  & 16.78  & \multirow{3}[2]{*}{14.06 } \\
          & ViT-Base & 8.43  & 47.08  & 21.73  & 5.38  & 20.66  &  \\
          & ViT-Large & 0.90  & 7.03  & 8.08  & 2.98  & 4.75  &  \\
    \midrule
    \multirow{3}[2]{*}{\textbf{DTA}} & ViT-Small & 92.56  & 99.43  & 98.82  & 81.46  & 93.07  & \multirow{3}[2]{*}{\textbf{85.52 }} \\
          & ViT-Base & 86.10  & 96.04  & 99.71  & 71.15  & 88.25  &  \\
          & ViT-Large & 71.54  & 92.01  & 96.32  & 41.12  & 75.25  &  \\
    \bottomrule
    \end{tabular}%
    \end{adjustbox}
  \label{tab:scale}%
\end{table}%

Here, we evaluate the three attacks on pre-trained models with varying scales. All models were pre-trained on ImageNet-1k using AugReg. As the results in Table \ref{tab:scale} show, our DTA beats the baseline methods by a considerable margin. Particularly, the average ASR of DTA is 85.52\%, which is 10.92\% and 71.46\% higher than NRDM and PAP, respectively. On ViT-Large, DTA achieves an average ASR of 75.25\% across the 4 datasets, which beats NRDM and PAP by 17.02\% and 70.5\%, respectively. Moreover, the sensitivity and instability of the baseline methods across different model scales are extremely high, which greatly limits their practicability when applied to diverse pre-trained models.

\subsection{Evaluation on Adversarially Pre-trained ViTs}

\begin{figure}
\centering
	\subcaptionbox{Full fine-tune}{\includegraphics[width = 0.23\textwidth]{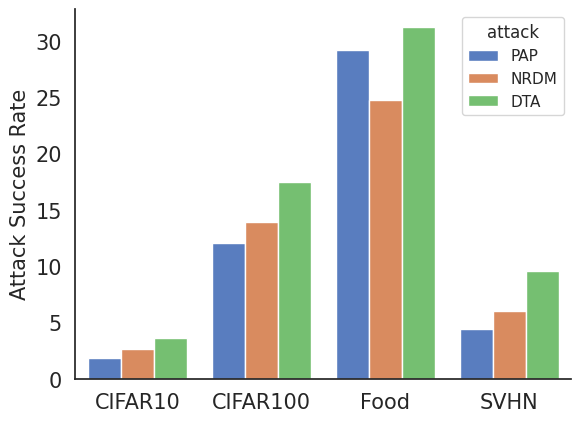}}
	\hfill
	\subcaptionbox{LoRA}{\includegraphics[width = 0.23\textwidth]{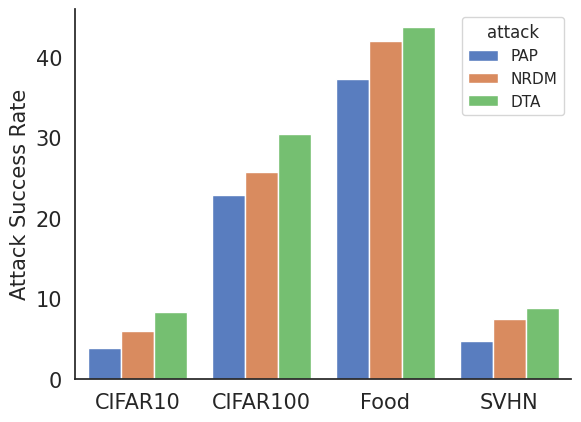}}
\caption{The ASR(\%) on adversarially pre-trained models. All downstream models are fine-tuned from the adversarial pre-trained XCiT-base model with $\epsilon = 8/255$ on ImageNet-1k.}

\label{fig:at}
\end{figure}

We also evaluate the attacks on adversarially pre-trained models, a common way to improve adversarial robustness. We take the XCiT model adversarially pre-trained (with $l_{\infty}$ budget $\epsilon = 8/255$) on ImageNet-1k by Edoardo et al.~\cite{debenedetti2023lightrecipe} as our pre-trained model. 
Due to the poor performance of AdaptFormer, here we mainly analyze full fine-tune and LoRA. As shown in  \Figref{fig:at}, unsurprisingly, the ASR decreases drastically for all attacks, which is well below 30\% and 45\% when transferred to attack the full or LoRA fine-tuned models, respectively. 
However, DTA is still comparably more effective than others. 
Another important observation is that although LoRA and full fine-tune have similar clean accuracy, the ASR on LoRA fine-tuned model is almost 10\% higher than that on full fine-tuned model. This uncovers a previously unknown weakness of LoRA: \emph{it is more vulnerable to downstream transfer attacks}.

\subsection{Attacking Object Detection and Segmentation}

\begin{table}[htbp]
  \centering
  \caption{Attacking object detection models on COCO2017.}
    \begin{tabular}{c|ccc}
    \toprule
    Attack & mAP   & mAP$_{50}$ & mAP$_{75}$ \\
    \midrule
    clean & 51.5  & 72.1  & 56.6  \\
    NRDM  & 45.8  & 66.0  & 49.8  \\
    PAP   & 35.7  & 54.3  & 38.3  \\
    \textbf{\textbf{DTA}}   & \textbf{35.6 } & \textbf{53.9 } & \textbf{38.0 } \\
    \bottomrule
    \end{tabular}%
  \label{tab:od}%

\end{table}%

\begin{table}[htbp]
  \centering
  \caption{Attacking segmentation models on ADE20k.}
    \begin{tabular}{c|cccc}
    \toprule
     & clean & NRDM  & PAP   & \textbf{DTA} \\
    \midrule
    mIoU  & 48.13 & 23.95 & 17.45 & \textbf{15.54} \\
    \bottomrule
    \end{tabular}%
  \label{tab:seg}%

\end{table}%

For object detection, we choose ViTDet~\cite{li2022vitdet} as the downstream model, which was fine-tuned on the COCO2017 dataset from an MAE pre-trained ViT-base model. Object detection tasks often pad different-sized images to the same size, it is thus meaningless to generate adversarial examples for the padding par. So we crop a 448x448 region from the upper-left corner of the image to perform the attack. As indicated in Table \ref{tab:od}, our DTA attack caused the model's mAPs to drop the most, from 51.5 to 35.6 on the COCO2017 validation set.

For the segmentation task, we take UPerNet~\cite{xiao2018upernet} as the downstream model, which was fine-tuned on the ADE20k dataset from an MAE pre-trained ViT-base model. We employ a 512x512 attack region here. As presented in Table \ref{tab:seg}, all attack methods result in a substantial decrease in mIoU, with our DTA being the most effective, reducing the mIoU to 15.54. This proves that our DTA can be generalized to attack different types of downstream tasks.

\subsection{Improving Adversarial Training}


\begin{table}[htbp]
  \centering
  \caption{The robustness (\%) of adversarially fine-tuned models by PAP and our DTA on CIFAR10.}
    \begin{adjustbox}{width=0.98\linewidth}
    \begin{tabular}{cccccc}
    \toprule
    Pretrain & Train$\downarrow$ Test$\rightarrow$ & DTA   & PAP   & Average & Clean Acc \\
    \midrule
    \multirow{2}[2]{*}{AugReg} & DTA   & 87.75 & 90.52 & \textbf{89.13} & 90.5 \\
          & PAP   & 26.25 & 97.89 & 62.07 & 97.69 \\
    \midrule
    \multirow{2}[2]{*}{MAE} & DTA   & 94.52 & 93.82 & 94.17 & 96.44 \\
          & PAP   & 96.25 & 97.1  & \textbf{96.67} & 96.45 \\
    \midrule
    \multirow{2}[2]{*}{DINO} & DTA   & 88.13 & 89.72 & \textbf{88.92} & 89.08 \\
          & PAP   & 24.35 & 98.36 & 61.35 & 98.28 \\
    \bottomrule
    \end{tabular}%
  \end{adjustbox}
  \label{tab:adv_ft}%
\end{table}%

Here, we show that DTA can help build better defenses against downstream transfer attacks. We follow the fine-tuning paradigm to finetune the downstream models on CIFAR10 using adversarial training~\cite{madry2017pgd}. 
The $\epsilon$ for adversarial finetuning is set to 4/255 to help maintain clean accuracy. But we use a larger $\epsilon = 10/255$ for testing. As shown in Table \ref{tab:adv_ft}, models fine-tuned with PAP can only defend PAP attacks and fail badly on DTA attacks, for AugReg and DINO pre-trained models. 
By contrast, using our DTA can help achieve universal robustness against different test attacks for all pre-trained models, resulting in significantly improved overall robustness. This highlights one unique benefit of sample-wise transfer attacks over UAPs.

\subsection{Ablation Studies}
We conduct 3 ablation studies on loss function, attack layers, and threshold $\gamma$. Here, we only report the main conclusions and defer the detailed results and analyses to appendix \ref{sec:ab}.

\paragraph{Loss Function }
We test 4 alternative loss functions for ATCS including the vanilla cosine similarity. Overall, ATCS is better than the vanilla cosine similarity in most cases and is more effective than other losses.

\paragraph{Attack Layer(s) }
We compare our DTA layer selection strategy with two alternative strategies and find that our strategy is better than attacking a fixed layer or all layers. 

\paragraph{Threshold $\gamma$}
We show that, as the threshold $\gamma$ increases from 0 to 1, the ASR first increases and then decreases, reaching its peak performance at $\gamma = 0.2$. And the trend is consistent across different downstream datasets.

\section{Conclusion}
\label{sec:conclusion}
In this paper, we studied the problem of downstream transfer attacks (DTAs) and explored how an attacker can generate highly transferable adversarial examples using a pre-trained model to attack downstream models fine-tuned by different techniques. We proposed to use Average Token Cosine Similarity (ATCS) as the adversarial objective and revealed that the ATCS value obtained at different layers is a good indicator of downstream transferability. With ATCS, we further proposed a DTA attack that can find the most vulnerable layer and generate highly transferable adversarial examples. Extensive experiments demonstrate the effectiveness of our DTA attack and its superiority over existing attacks. We also found that emerging PETL methods like LoRA are more susceptible to transfer attacks crafted on the pre-trained model.
We also show that our DTA can help train more robust models resistant to downstream transfer attacks when applied with adversarial training.

\clearpage
\bibliography{main}
\bibliographystyle{icml2024}
\clearpage
\setcounter{section}{0}
\renewcommand\thesection{\Alph{section}}

\section{Clean Accuracy of Downstream Models}
Table \ref{tab:clean} reports the clean accuracies of the victim downstream models. As can be observed, the finetuned models all perform well on the clean datasets, regardless of the finetuning method.

\begin{table*}[htbp]
  \centering
  \caption{The clean accuracy (\%) of the fine-tuned models with different fine-tune and pre-train methods. All models are ViT-base and the pre-training was done on ImageNet-1K.}
  \begin{adjustbox}{width=0.98\linewidth}
    \begin{tabular}{cccccccccccc}
    \toprule
    Fine-tune & Pre-train & CIFAR10 & CIFAR100 & STL10 & Cars  & Cub   & DTD   & FGVC  & Food  & Pets  & SVHN \\
    \midrule
    \multirow{3}[2]{*}{Full } & MAE   & 98.22 & 82.11 & 97.43 & 84.44 & 74.66 & 65.21 & 48.24 & 85.71 & 91.66 & 96.67 \\
          & DINO  & 98.77 & 90.05 & 99.09 & 90.19 & 83.86 & 74.79 & 65.86 & 89.97 & 93.38 & 97.3 \\
          & AugReg & 98.62 & 88.53 & 98.8  & 85.97 & 83.39 & 71.75 & 60.07 & 88.1  & 94.03 & 96.83 \\
    \midrule
    \multirow{3}[2]{*}{LoRA} & MAE   & 97.83 & 83.55 & 97.86 & 88.99 & 80.68 & 69.3  & 67.33 & 87.34 & 93.1  & 97.44 \\
          & DINO  & 98.62 & 88.54 & 98.98 & 87.85 & 79.85 & 72.23 & 65.47 & 87.2  & 92.75 & 97.04 \\
          & AugReg & 98.24 & 87.4  & 98.73 & 82.35 & 74.24 & 69.46 & 55.9  & 86.14 & 92.12 & 97.25 \\
    \midrule
    \multirow{3}[2]{*}{AdaptFormer} & MAE   & 97.62 & 80.87 & 97.62 & 87.87 & 77.45 & 68.61 & 64.63 & 87.51 & 91.96 & 97.33 \\
          & DINO  & 98.29 & 85.09 & 98.66 & 82.41 & 76.45 & 71.11 & 59.44 & 86.79 & 92.56 & 96.27 \\
          & AugReg & 98.01 & 84.6  & 98.7  & 76.92 & 70.33 & 67.92 & 49.61 & 84.24 & 91.63 & 96.75 \\
    \bottomrule
    \end{tabular}%
  \end{adjustbox}
  \label{tab:clean}%
\end{table*}%

\section{Ablation Study}
\label{sec:ab}
\begin{figure*}
\centering
	\subcaptionbox{AugReg}{\includegraphics[width = 0.33\textwidth]{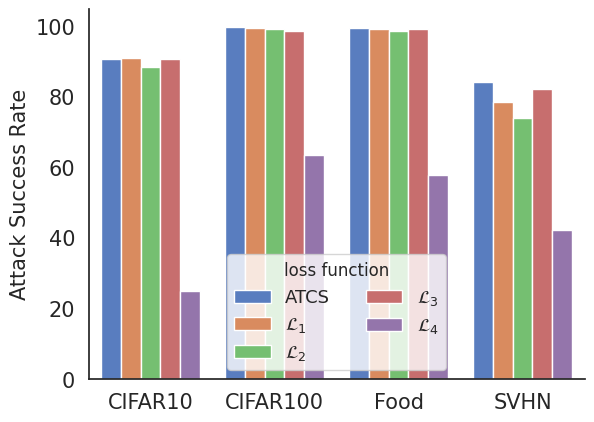}}
	\hfill
	\subcaptionbox{DINO}{\includegraphics[width = 0.33\textwidth]{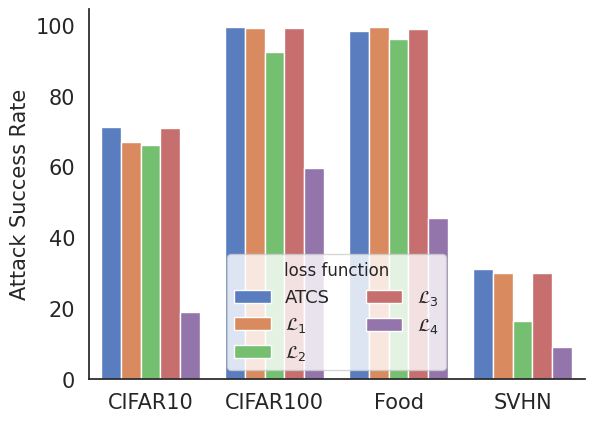}}
	\hfill
	\subcaptionbox{MAE}{\includegraphics[width = 0.33\textwidth]{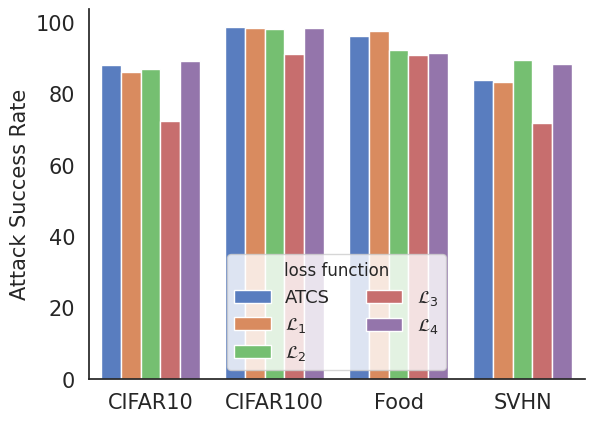}}

\caption{The ASR(\%) of different loss functions on AugReg, DINO, MAE pre-trained models. The attack layer was set to 1, 8, and 8 for MAE, DINO, and AugReg, respectively.}
\label{fig:loss}
\end{figure*}

\begin{figure}[htbp]
    \centering
    \begin{minipage}[t]{0.23\textwidth}
        \centering
      \includegraphics[width=1\textwidth]{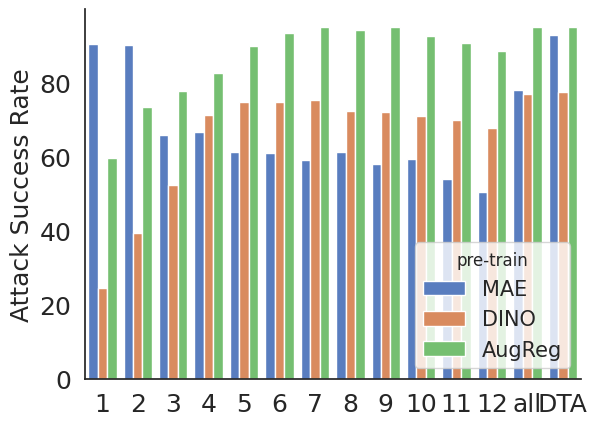} 
      \caption{Comparing different layer selection strategies. $x/y$-axis: attacking layer/ASR.} 
      \label{fig:layer} 
    \end{minipage}
    \hfill
    \begin{minipage}[t]{0.23\textwidth}
        \centering
      \includegraphics[width=1\textwidth]{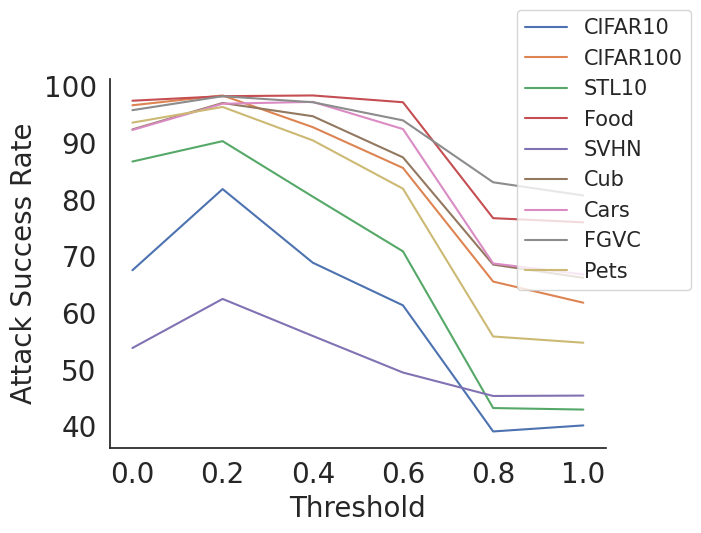} 
      \caption{The ASR (\%) with different threshold $\gamma$, averaged over the 3 pre-training methods.} 
      \label{fig:threshold} 
    \end{minipage}
\end{figure}





Here, we analyze the impact of the adversarial loss function and the two hyper-parameters including layer $k$ and threshold $\gamma$ to DTA. We fix the fine-tuning method to full fine-tune and set the attack step size to 0.03 and step number to 10.

\paragraph{Loss Function }
Here, we test 4 alternative loss functions for our ATCS:\\
$\Ls_1 = - cos(f^k(x),f^k(x'))$ \quad $\Ls_2 = \norm{f^k(x)-f^k(x')}$ \\
$\Ls_3 = - f^k(x) \cdot f^k(x')$ \quad\quad\quad $\Ls_4 = \norm{f^k(x')}$ \\
As illustrated in \Figref{fig:loss}, although $\Ls_4$ has been proven to be effective in the UAP setting~\cite{ban2022pap, mopuri2017uap-fastfeaturefool}, it does not perform well (or at least not as well as other losses) in the downstream transfer setting, which is a sample-wise attack scenario. Overall, our ATCS achieves a slightly better performance than the vanilla cosine similarity in most cases and is more effective than other loss functions.

\paragraph{Attack Layer(s) }
Here, we compare our DTA layer selection strategy with several alternative layer selection strategies, such as selecting a particular layer or all layers together. We report the average ASR of attacking 10 downstream models with different pre-train methods. As shown in \Figref{fig:layer}, our DTA achieves a higher ASR than attacking a particular layer or all layers. For MAE pre-trained models, attacking the shallow layer is better, while for DINO and AugReg pre-trained models, attacking the middle layer is better. Attacking all layers works well for DINO and AugReg pre-trained models with comparable performance to DTA, but is inferior to DTA for MAE pre-trained model. Overall, our DTA works the best among the three layer selection strategies.

\paragraph{Threshold }
Here we study the effect of threshold $\gamma$. \Figref{fig:threshold} shows that, for various downstream datasets, as the threshold value increases from 0 to 1, the ASR first increases and then decreases, reaching its peak performance at $\gamma = 0.2$. And the trends are quite consistent on different downstream datasets. Arguably, the optimal $\gamma$ may vary in real-world scenarios, which can be carefully tuned if the attacker knows more information about the downstream task.

\section{More Understandings of DTA}
\begin{figure}
\centering
\includegraphics[width=.4\textwidth]{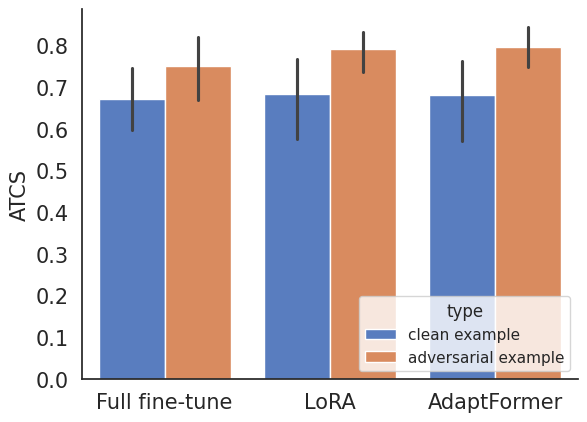}
\caption{ATCS between the pre-trained model and the downstream model. The blue bar represents the ATCS of $f_\theta^{k}(x)$ and $f_{\theta'}^{k}(x)$, while the yellow bar represents the ATCS of $f_\theta^{k}(x')$ and $f_{\theta'}^{k}(x')$. We set the attack layer $k=8$.}
\label{fig:cos_transfer}
\end{figure}

Undoubtedly, representation/feature reuse is one key aspect of the pretraining-and-finetuning paradigm. The downstream model inherits the internal loss landscape and features of the pre-trained model~\cite{neyshabur2020What-is-being-transferred}. This implies that the perturbations generated to distort the pre-trained features are to some extent also disruptive to the downstream model.
To better understand this, we feed the clean and adversarial examples separately into the pre-trained vs. fine-tuned models and compare the obtained ATCS, i.e., $\text{ATCS}(f_\theta^{k}(x'), f_{\theta'}^{k}(x'))$ and $\text{ATCS}(f_\theta^{k}(x), f_{\theta'}^{k}(x))$. Here, we set the attack layer to $k=8$. As shown in \Figref{fig:cos_transfer}, the ATCS of the adversarial examples is higher than that of the clean samples, which implies that the vulnerabilities of the pre-trained model explored by the adversarial samples are preserved in the fine-tuned model.
Notably, when fine-tuned using PETL methods, the downstream model becomes more similar to the original model in the feature space. This similarity also predicts that PETL models are more susceptible to downstream transfer attacks.


There is also an interesting shift of focus on the attacking layers targeted in different works, i.e., earlier works often attack the middle layers~\cite{lu2020enhancing, naseer2020self, zhang2022BIA}, while more recent works favor the shallow layers~\cite{ban2022pap, zhang2022BIA}. Based on our analyses and empirical observations, this might be related to the trend that earlier models pre-trained using supervised learning are more vulnerable in the middle and deep layers, while more recent models pre-trained by self-supervised learning are more vulnerable in the shallow layers. Our DTA provides a simple but effective technique to explore the most vulnerable layers of the pre-trained model and thus can be effective for different pre-training methods.

\end{document}